\documentclass[10pt]{article}

\usepackage{amsmath}
\usepackage{amsthm}
\usepackage{amsfonts}
\usepackage{amssymb}
\usepackage{lipsum}
\usepackage{blindtext} 
\usepackage{mathtools}
\usepackage{thmtools}
\usepackage{enumitem}
\usepackage{xcolor}
\usepackage{cite}
\usepackage{mathtools}
\mathtoolsset{showonlyrefs=true}
\usepackage[title]{appendix}

\usepackage{braket}
\usepackage{cancel}

\usepackage{graphicx}
\usepackage{floatrow}
\begin{document}
\date{}
\author{G. Citti, A. Sarti
\footnote{The authors have been supported by Horizon 2020 Project ref. 777822: GHAIA }
} 
\title{Neurogeometry of perception:\\ isotropic and anisotropic aspects
}

\maketitle


\section{Introduction}

The first attempts to formalize rules of perception go back to the 
Gestalt psychologists, as for example  Wertheimer \cite{Wertheimer},  Kohler \cite{Kohler1929}, Koffka \cite{Koffka1935}. They formulated some geometric laws accounting for perceptual phenomena, taking into account different features like  position but also  brightness, orientation and scale. More precisely these laws depend on normalized differences of position, brightness, orientation or scale between the target image and the background. As an example we recall that two circles, with 
equal luminance, (so that they reflect physically the same
amount of light) are perceived of different grey level, if put on different backgrounds: 
the one with darker background is perceived as 
brighter than the one with bright background (see figure 1). This phenomenon, 
also called simultaneous contrast, was 
known by Goethe, and then it  has been deeply studied. 
A first explanation is that the brightness is related 
to detection of edges of the region, followed by a filling-in 
mechanism (\cite{grossberg1988Todorovic}, \cite{Rossi1996},  \cite{ShapleyEnroth}). 
In particular the retinex model was inspired by similar principles 
\cite{BrainardWandell, McCann, Hurlbert, Kimmel, Morel}.
These aspects have to be considered as relatively
local properties of perception. A number of computational models have been published, taking into account other aspects of perception as for example distal aspects, 
or illumination  (see for example \cite{ArendBuehler}, 
 \cite{Blakeslee}, and \cite{Todorovic}).

\begin{figure}[H]\label{fig1}
\centering
\includegraphics[height = 2 cm]{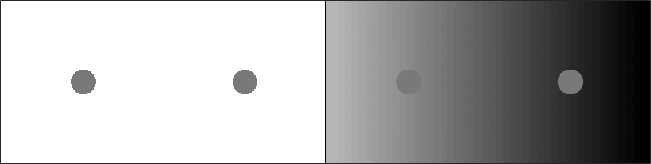}
\;\;\;\;\;\;\;\;\;
\includegraphics[width = 3 cm]{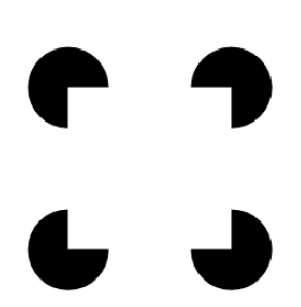}
\caption{Left: Two circles with the same intensity of grey. Middle: the same circles are represented with a non constant background. As a result the circle on the right appears to be brighter than the other one which has a darker background. Right: The Kanizsa square.}
\end{figure}

Here we will consider an other aspect of perception: its globality, since 
emergence of perceptual units depends on the visual stimulus as a whole. Indeed visual scenes  are  perceived  as  constituted  by  a  finite  number  of  figures.  The  most  salient  configuration  pops  up  from  the ground  and  becomes  a  figure  (see Merleau-Ponty  \cite{Merleau-Ponty1945}). Often this happens  even in absence of gradients or boundaries in the visual stimulus (see for example the classical example of the Kanizsa triangle \cite{Kanizsa1979}). A long history of phenomenology of perception, starting from Brentano, to Husserl until Merleau-Ponty has considered the emergence of foreground and background at the centre of perceptual experience and resulting as an articulation of a global perceptual field.

A number of researchers studied principles of psychology of form  in this perspective not only qualitatively but also quantitatively. We recall the first results of  Grossberg and Mingolla in \cite{grossberg1985neural}, the notion of vision field proposed by Parent and Zucker in  \cite{parent1989trace} and the  theory of object perception due to Kellman and Shipley in \cite{kellman1991theory} \cite{shipley1992perception,shipley1994spatiotemporal}. In the same years Field, Hayes and Hess \cite{field1993contour} introduced the notion of association fields, via the following experiment. figure 2(a)  it composed by random Gabor patches and a perceptual units obtained via aligned patches, which is 
also visualized in figure 2(b). Through a series of similar experiments, Field, Hayes and Hess  constructed an association field, which describes the complete set of possible subjective boundaries starting from a fixed point. See figure 2(c).

\begin{figure}[H]\label{fig_association}
\centering
\includegraphics[height = 2.8 cm]{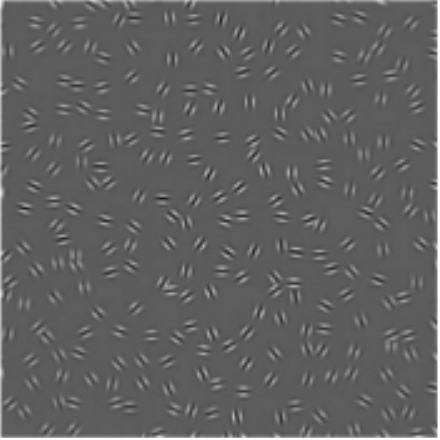}
\includegraphics[height = 2.8 cm]{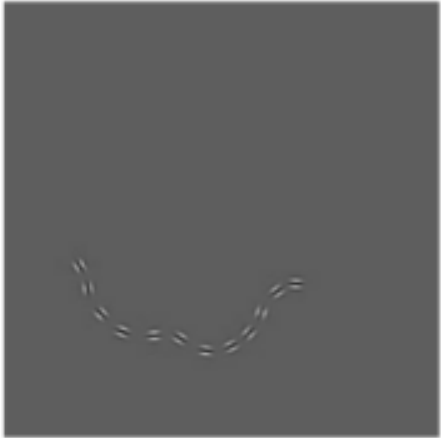}
\includegraphics[height = 1.8 cm]{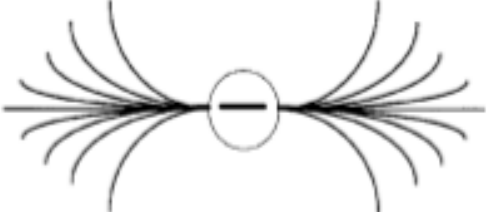}
\caption{Left: The image proposed in the experiment of Field, Heyes and Hess \cite{field1993contour}. Middle: The perceptual unit present in the image on the left. Right: The association field.}
\end{figure}

\bigskip

We take the point of view that the geometry of visual perception is induced by the functional geometry of the visual cortex.
The first results in this direction have been achieved by 
Hubel and Wiesel \cite{hubel1977ferrier} who provided experimental evidence of the
presence in the primary cortex of families of cells sensitive to 
different orientations organized in the so called hypercolumnar module.
After that, results of Bosking et al. in \cite{bosking1997orientation}
and Fregnac  Shulz in \cite{fregnac1999activity} provided experimental evidence for the organisation of many other features already studied in psychology of perception, like brightness, curvature and scale. 
In addition, the structure of horizontal connectivity between cells studied by Bosking 
is a good candidate as a neural correlate for perceptual association fields. 
This allows to introduce a relation between geometry of 
the cortex and geometry of vision. Mathematical models in this 
direction have been proposed by Hoffman in \cite{Hoffman}, 
 Mumford in \cite{mumford1994elastica}, Williams and Jacobs in \cite{williams1997stochastic} and August Zucker in \cite{august2000curve}. They modelled the analogous of the association fields with Fokker-Planck equations. Petitot and Tondut (in \cite{PetitotTondut}) introduced a model of the functional architecture of V1, and described the propagation by a constrained Lagrangian operator, giving a possible explanation of Kanizsa subjective boundaries in terms of geodesics.
With these results it became clear the central role of geometry 
in the description of the organization of the cortex and of its correlate on the 
 visual plane. 
 Citti and Sarti in \cite{CittiSarti} proposed a model of the functional architecture of V1 as a Lie group of symmetry, with a sub-Riemannian geometry, showing the strict relation between geometric integral curves, association fields, and metric cortical properties. 
A large literature has been provided on models  in the  same subriemannian  space both for boundary  completion 
\cite{BenShaharZucker, GuyShahar, RenMalik, Duits1,  Duits2, Duits} and for other perceptual phenomena \cite{sarticittinoncommutative, Functional, Franceschiello}. 
 This approach has been called neurogeometry (or neuromathematics).
For a more complete bibliography and application fields of neuromathematics the reader can refer for example to \cite{CittiSartibook}. 
 \bigskip

These models assume by simplicity that the perception is orientation 
invariant, allowing a representation in the Lie group of rotation and translation. 
However it is a 
classical finding that human subjects perform best on a
spatial acuity test when the visual targets are oriented horizontally
or vertically. 
This phenomenon first described 
by Mach in \cite{mach}, has been called the oblique effect by Appelle \cite{appelle}. It applies to different phenomena such as orientation selectivity
(Andrews \cite{Andrews}; 
Campbell and Kulikowski \cite{Campbell}; Orban et al.\cite{Orban}), 
orientation
discrimination (Bouma and Andriessen \cite{Bouma}; Caelli et al.
\cite{Caelli}, Regan and Price \cite{Regan}; Westheimer and Beard \cite{Westheimer}), 
grouping (Beck \cite{BECK1, BECK2, BECK3}), 
and geometric illusions (Green
and Hoyle \cite{GREEN} Leibowitz and Toffey
\cite{Leibowitz}, Wallace \cite{Wallace}). In particular in 
Bross et al. \cite{Bross} the oblique effect is studied on a Kanizsa figure. A Kanizsa square and a Kanizsa diamond are considered. If no edge misalignment is present both the square and the diamond are correctly reconstructed. However with a 6-degree misalignment the square is not perceived, while the diamond is perceived. For higher misalignment neither the square not the diamond are perceived (see figure 3).

\begin{figure}[H]
\begin{center}\label{Kan}
\includegraphics[height=4 cm]{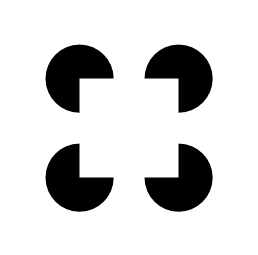}
\includegraphics[height=4 cm]{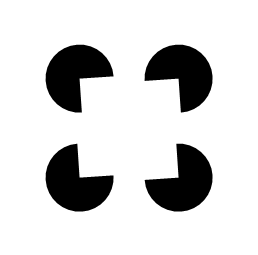}
\includegraphics[height=4 cm]{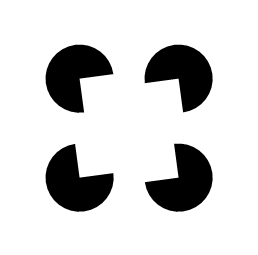}\\
\includegraphics[height=4 cm]{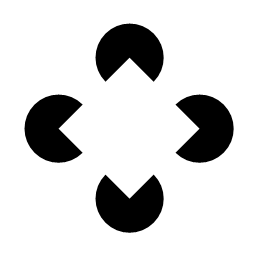}
\includegraphics[height=4 cm]{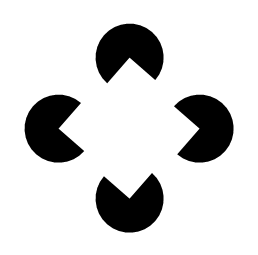}
\includegraphics[height=4 cm]{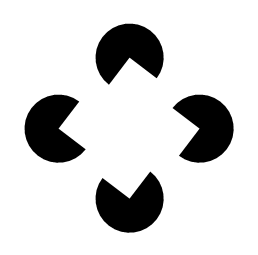}
\end{center}
\caption{Kanizsa squares and diamonds with progressively more misaligned edges. From  left to right the misalignment is 0, 6 and 12 degrees. }
\end{figure}

Neural correlates  of this phenomenon have been provided by Hubel and Wiesel  \cite{hubel1962receptive}, who proved that the distribution of oriented cells 
who respond to lines is not uniform. 
Nowadays most studies agree that the oblique effect happens 
in the visual cortex \cite{cinque} \cite{sei} but the mechanism is not totally clear.  However it seems that there are at least two aspects 
to be considered: the functionality of the visual system 
(see   \cite{uno} and \cite{Aertsen}), 
and  learning from the environment  (e.g. with statistics of natural scenes - 
see    \cite{dodici} and \cite{Essock}).

In this paper we first recall the definition of neurogeometical model for a general set of features, focusing in particular  on the geometrical properties of horizontal cortical 
connectivity and showing how they can be considered as a neural correlate of a geometry 
of the visual plane. 
In this first geometrical model 
the geometry of the cortex is expressed as a sub-Riemannian 
geometry in the Lie group of symmetries of the set of features. 

Here, using data taken from \cite{sanguinetti} 
we recognize that histograms of edges - co-occurrences 
are not isotropic distributed, and are strongly biased 
in horizontal and vertical directions of the stimulus.
Finally we introduce a new model of non isotropic 
cortical connectivity modeled on the histogram of 
 edges - co-occurrences. 
Using this kernel in the geometrical cortical 
model previously described, we are able to justify 
oblique phenomena comparable with the 
experimental findings of \cite{Bouma} and \cite{Bross}. 

\section{A General Approach to the Geometry of Perception}

 Citti and Sarti developed in a few papers \cite{CittiSarti, 
 SartiCittiPetitot, sarti2015constitution, gaugecortex}, a theory of invariant perception in  
Lie groups, taking into account different features: brightness
orientation, scale, curvature, movement. Each of these 
features have been characterized by a different Lie group, but 
the different techniques can be presented from an unitary point of view, which we recall here.

\subsection{Families of cells and their functionality}

The primary visual cortex is the first part of the brain processing the visual signal coming from the retina. The receptive field (RF) of a cortical neuron is the portion of the retina which the neuron reacts to, and the receptive profile (RP)  $\psi({ \xi})$ is the function that models the activation of a cortical neuron when a stimulus is applied to a point 
${\xi} = (\xi_1,\xi_2)$ of the retinal plane. 
 
Due to the retinotopic structure, there is an log-polar map between retina and cortical space in V1, 
which we will discard in first approximation. 
In addition the hypercolumnar structure organizes the cortical cells of V1 in
columns corresponding to different features.
As a results 
we will identify cells in the cortex by means of 
three parameters $(x_1,x_2,f)$, where $(x_1,x_2)$ is 
the position of the point, and $f$ a vector of extracted features. We will denote $F$ the set of features, and consequently 
the cortical space will be identified with $R^2 \times F$.  
In the presence of a visual stimulus the whole hypercolumn fires, giving rise to an output 
\begin{equation}\label{output_general}
O_{F}(x_1,  x_2, f) = \int I(\xi_1, \xi_2) \psi_{(x_1, x_2, f)} (\xi_1, \xi_2)d\xi_1 d\xi_2.
\end{equation}
Note that the output is a high dimensional function, 
defined on the cortical space. 
We denoted it $O_{F}$ to underline the dependence on the family of filters $F$. It is clear that the same image, filtered with a different family of cells, produces a different output. 

\subsection{Cortical Connectivity and the Geometry of the Cortical Space} 

The output of a family of cells is propagated in the cortical space 
$R^2 \times F$ via the lateral connectivity. 
We will represent connectivity as a geometric kernel $K_F$, acting in the cortical space 
$R^2 \times F$ on $O_{F}$ which is now considered a feedforward input to the overall activity. The shape of this kernel 
will be different in different families of cells and will be compatible with their invariance properties. We will see that there 
is a strict relation between the structure of set $R^2 \times F$ of receptive profiles, their functionality 
and the direction of propagation of cortical activity
codified in the kernels $K_F$. 
This allows to introduce a geometry of the 
considered family of cells, where the distance is 
not the usual Euclidean one, but it is induced by the 
strength of connectivity: the distance between couple 
of points will be considered a decreasing function of the 
value of the connectivity strength.

Different equations associated to these kernels have been used 
to describe the propagation of neural activity  in the cortical space 
$R^2 \times F$. 
In particular, following an approach first proposed by 
  Ermentraut-Cowan in \cite{ErmentroutCowan} and Bressloff and Cowan 
in \cite{bressloff2003functional}, 
we consider a mean field equation representing the cortical activity $a( x_1, x_2, f)$. In the stationary case the equation becomes
\begin{equation}
\label{mean_field}
 a( x_1, x_2, f) = \sigma\Bigg(\int K_F((x_1,x_2,  f , \bar x_1, \bar x_2, \bar f ) \Big(\frac{a(\bar x_1, \bar x_2, \bar f) + 
O_{F}(\bar x_1, \bar x_2, \bar f )}{2}\Big)d\bar x_1 d\bar x_2 d\bar f  \Bigg), 
\end{equation}
where $\sigma$ is a sigmoidal functional.
Our intent in the following is to show that the RP functions $\psi$ as well as the connectivity kernels  $K_F$ can be derived from theoretical geometrical considerations.

\subsection{Spectral analysis and perceptual units} 
Bressloff, Cowan, Golubitsky and Thomas  in \cite{bressloff2002geometric}, 
showed that in absence of a visual input $O_{F}$  but in presence of drugs excitation, the 
eigenvectors of this functional on the whole cortical 
space can describe visual hallucination phenomena. 

In presence of a visual input, the propagation of the signal through 
horizontal connectivity and its eigenvectors have 
been studied by 	Sarti and Citti in \cite{sarti2015constitution}.
Since eigenvectors are 
the  emergent  states  of  the  cortical  activity,  they individuate  the  coherent  perceptual units  in  the  scene  and  allow  to  segment  it.  Every eigenvector corresponds to a different perceptual unit. 
Let  $\Omega_F$ be the set where the feedforward input is bigger than a fixed threshold $h$:
\begin{equation}\label{spectral}
\Omega_{F}= \{(x_1,  x_2, f): |O_{F}(x_1,  x_2, f)|\geq h \}.
\end{equation}
Calling 
$K_F * v$
the restriction to $\Omega_F$ of the 
linearization of the operator defined in \eqref{mean_field} we get up to a change of variable the operator: 
\begin{equation}
(K_F * v_i) (x_1,x_2,  f)  =  \int K_F((x_1,x_2,  f) ,( \bar x_1, \bar x_2, \bar f )) v_i(\bar x_1, \bar x_2, \bar f) d\bar x_1 d\bar x_2 d\bar f  
\end{equation}
The	associated eigenvectors 
satisfy the equation: 
\begin{equation}\label{spectral_general}
K_F * v_i  = \lambda_i v_i 
\end{equation}
and correspond to visual units (see \cite{sarti2015constitution} for details).
This result shows that perceptual units correspond to visual hallucination modulated by the stimulus.
Eigenvectors of these operators are functions defined on 
$\Omega_F$ with real values.  We  will  assign 
the meaning of a saliency index of the objects to the associated eigenvalues. 
The  first  eigenvalue  will  correspond  to the  most  salient  object  in  the  image. In addition, the associated 
eigenvalue allows comparison between different scenes and images. 
This approach  
	also extends on neural basis the model proposed for  image  processing   by Perona   Freeman,  \cite{perona1998factorization},  Shi   Malik,  \cite{shi2000normalized},  Weiss \cite{weiss1999segmentation},  Coifman   Lafon
	 \cite{coifman2006diffusion}.

\subsection{Projected geometry on the visual plane} 

The geometry of visual perception is simply the projection of the active
cortical geometry on the visual plane. 
Even though it is not clear where this phenomenon is 
codified we will project on the retinal plane the geometry previously introduced in 
order to describe geometry of vision. 

For every cortical point, the cortical activity, suitably normalized can be considered a probability density. 
Hence its maximum over the fibre $F$ can be considered the most probable 
value of $f$, and can be considered the feature identified by the system: 
\begin{equation}\label{max_general}
|O_{F}(x_1, x_2,  f_I(x_1, x_2))| = \max _f |O_{F}(x_1, x_2, f)|.
\end{equation}

Note that the function $f_I$ detect at every point the most 
salient value of the feature, and clearly depends on the properties of the visual stimulus $I$. This feature selection mechanism induces a geometry on the perceptual plane. 
The projection of the kernel $K_F$ on the perceptual plane 
will be a kernel $K_{\Pi}$ defined as

\begin{equation}
\label{kernel_projection}
K_{\Pi}( (\xi_1, \xi_2), (x_1, x_2)) = K_F(
(\xi_1,\xi_2, f_I (\xi_1,\xi_2)), (x_1, x_2,  f_I( x_1, x_2))). 
\end{equation}

Note that the  kernel $K_{\Pi}$ 
is now depending on the function $I$. 
Hence it carries the metric induced on the retinal  plane by the image $I$. 

If we propagate the function $I$ with this kernel, we obtain 
an activity function on the retinal plane
\begin{equation}\label{activ_general}
u(\xi_1, \xi_2) = \sigma\Bigg(\int K_{\Pi}((\xi_1, \xi_2), ( x_1, x_2))  O_{ F}(x_1, x_2, f_I(x_1, x_2) ) d x_1 d x_2 \Bigg). 
\end{equation}

In the following we will discard 
the sigmoid $\sigma$ when the solution 
remains bounded from above and below.

\subsection{The modular structure of the cortex} 

We can assume that different families of cells 
depending on different features 
act in sequence on the same visual stimulus. 
A family of cells 
will process the stimulus in a 
space $R^2\times F_1$. Then its output will become the 
input for the second family of cells, 
(depending for example on orientation) which will process 
the stimulus in a new space of features  $R^2\times F_2$. 
In this way interaction between the two 
families of cells can be described with our model, 
 even if they are selective for 
different features.

\section{Isotropic models inherited by the geometry of the visual cortex}

\subsection{Brightness perception}

{We refer to the introduction and to \cite{Todorovic} and the references therein for a review on the state of the art. In our model the visual signal is convolved with the RF of the retinal cells, which are modeled as a Laplacian of a Gaussian. After that we apply a distal mechanism induced by the non classical components of RF. Indeed in Lateral Geniculate Nucleus (LGN) e in V1 there is a strong near surround suppression mechanism due to the lateral connection  and a far surround suppression mechanism due to feedback connectivity from higher cortices \cite{Angelucci}. We postulate that the far surround suppression mechanism is able to generates brightness contextual effects and that it is at the origin of filling in.  Indeed the simplest surround function with positive values at the origin and negative at infinity is a negative logaritmic fuction, which coincides with the fundamental solution of the Laplace operator. As a result the output of the mexical hat RPs is convolved with a logaritmic functioin, and provides the reconstruction of the signal up to an harmonic function, which takes into account a global light contrast effect.}

 \subsubsection{The set of receptive profiles}
Retinal ganglion cells and LGN cells, have a radially symmetric shape (see \cite{DeAngelis}), and are ubbsually modeled by Laplacians of Gaussians:

$$\psi_{0, LGN}(\xi_1,\xi_2)= \Delta G(\xi_1,\xi_2)$$

where $G(\xi_1,\xi_2)= e^{-(\xi_1 ^{2}+\xi_2^{2})}$ is the Gaussian bell.
We will consider by simplicity filters at a fixed scale (see figure 4).
As a result there is a single receptive 
field over each retinal point $(\xi_1,\xi_2)$. 
In other words all filters of this family of cells 
can be obtained from a fixed one by translation: 
$$\psi_{x_1,x_2, LGN}(\xi_1, \xi_2)= \psi_{0, LGN}(x_1-\xi_1, x_2-\xi_2).$$
Formally the set of filters will be identified with a copy of the group $R^2$, with the standard sum. 
\begin{figure}
\begin{center}
\includegraphics[height=1.4in]{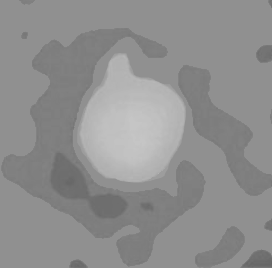}
\includegraphics[height=1.4in]{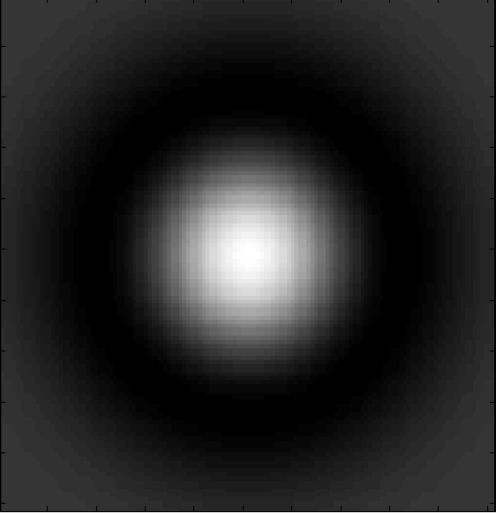}
\caption{ Left: Measured LGN cell RP  from  De Angelis \cite{DeAngelis}.Right: A model with a Laplacian of a Gaussian.
}
\end{center}
\end{figure}

In presence of a visual stimulus $I$, the retinotopic map performs a 
logarithmic deformation, which will be discarded here for simplicity.  Then the 
output defined in \eqref{output_general} reduces to 
$$
O_{LGN}(x_1,x_2)=\int I(\xi_1,\xi_2) \psi_{x_1, x_2, LGN}(\xi_1, \xi_2) d\xi_1 d\xi_2 = \quad \quad \quad  $$$$
\int  I(\xi_1,\xi_2)\psi_{0, LGN}(x_1- \xi_1, x_2- \xi_2)  d\xi_1 d\xi_2
=
(\Delta G*   I)(x_1,x_2)\approx \Delta I(x_1,x_2). 
$$

 \subsubsection{Connectivity patterns and the Euclidean  geometry of the space}
The output of LGN cells is propagated via the lateral connectivity. 
Experimentally it has been showed that connectivity is isotropic. 
As a result both the filters and the connectivity has the same radial symmetry. This is typical of isotropic kernels of the Euclidean geometry, 
and it is compatible with the fact that 
the set of cells is simply defined on the $R^2$ plane. 
Hence we can choose as model of connectivity the fundamental solution of the 
Laplace operator in the isotropic Euclidean metric: 
$$K_{LGN}(x_1,x_2) = \log(x_1^2 + x_2^2).$$
According to \eqref{activ_general} total activity contribution will be modelled as 
$$u(\xi_1, \xi_2)  = \int K_{LGN}(\xi_1 - x_1, \xi_2-x_2)  O_{LGN}(x_1,x_2) dx_1dx_2$$
Note that we are identifying the set of filters with $R^2$, the geometry 
is the Euclidean one and the propagation is performed 
with the standard Laplacian on it. This does not means that 
the perceived image coincide in general with the visual input. Indeed 
$O_{LGN}(x_1,x_2) = \Delta I$, which implies that $u - I$ is an harmonic function. If we impose Neumann boundary conditions,  the same harmonicity condition can be expressed as minimum of the functional 
\begin{equation}\label{functional1}
L_1(u) = \int |\nabla (I-u)|^2 dx_1dx_2 .
\end{equation}
Finally we propose to obtain the brightness of the 
perceived targets as a mean between the initial image and the activity $u$:
\begin{equation}\label{br}b=(I+u)/2.\end{equation}
Also note that the perception of the background is unchanged. 

We apply this algorithm to the image in figure 1 left and the results are shown in figure 5.
On the left it is represented the output $O_{LGN}$, which is a convolution with a Laplacian of a Gaussian. Then we compute the activity:
due to the Neumann boundary condition the background has a constant color, 
and the circle on the right has gray level brighter than the one on the left. 
Finally, applying equation \eqref{br}, we obtain the perceived gray level of the two target circles 
 as a mean of the stimulus and the activity (figure 5 right). 

\begin{figure}[H]\label{reti}
\begin{center}
\includegraphics[height= .9 in]{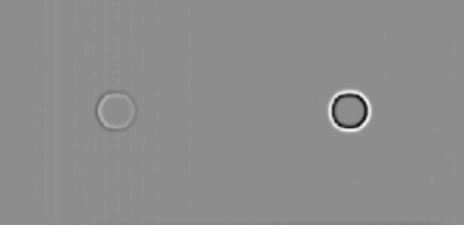}
\includegraphics[height= .9 in]{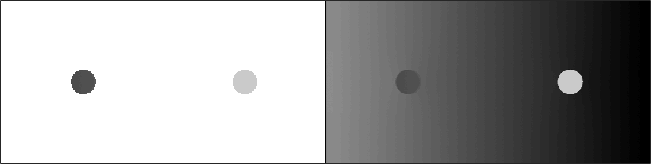}
\caption{We apply our model to the image of figure 1. Left: the output $O_{LGN}$ computed  as convolution with Laplacian of a Gaussian. Middle: after applying our algorithm to figure 1, we depict here the processed circles. Right: the reconstruction operated by our algorithm: 
the circle on the right appears brighter than the other, since it surrounded by a darker background.}
\end{center}
\end{figure}

\subsection{Geometry of Orientation space}

\subsubsection{The family of simple cells as the Lie group of rotation and translation}
Receptive profiles of simple cells sensitive to orientation have been modeled as Gabor filters by Daugman \cite{daugman1985uncertainty}, Jones and Palmer  \cite{jones1987evaluation}.
A suitable choice of the mother function is $$\psi_{0, V1}(\xi_1, \xi_2) = e^{- i \xi_2 -(\xi_1^2 + \xi_2^2)}.$$
The previously mentioned authors also showed that every simple cell is obtained from the mother one via rotation, translation. If we denote 
 $T_{x_1, x_2}$  a translation of a vector 
$(x_1, x_2)$ and $R_\theta$ a rotation of an angle $\theta$, their expression becomes: 
$$\psi_{x_1, x_2, \theta, V1}(\xi_1,\xi_2) = \psi_{0,  V1}(T_{x_1, x_2}R_ \theta)(\xi_1,\xi_2))= $$
\begin{equation}\label{filtri}=
e^{- i \Big((\xi_1-x_1) \sin(\theta) - (\xi_2-x_2) \cos(\theta)\Big) -((\xi_1-x_1)^2 + (\xi_2-x_2)^2)}.
\end{equation}

In this way any simple cell receptive profile is identified 
by an element of the Lie group of rotation and translation 
$R^2 \times S^1$ (see fig. 6 left and middle).

\begin{figure}\label{gabor_conn}
\begin{center}
\includegraphics[height=3 cm]{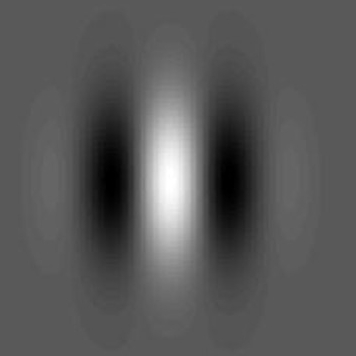}
\includegraphics[height=3 cm]{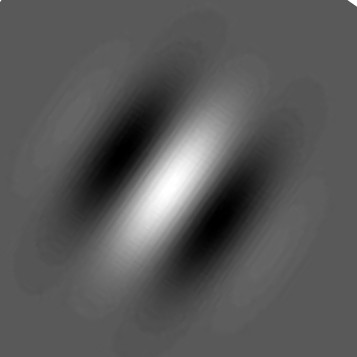}
 \includegraphics[height=3 cm]{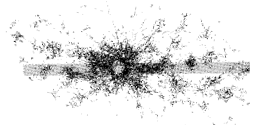}
 \caption{Left and Middle: two Gabor filters with different orientation. 
 Right: the connectivity pattern, measured by Bosking \cite{bosking1997orientation}}
\end{center}
\end{figure}

The output of this family of cells receptive profiles 
can be expressed by linear filtering of the stimulus $I$ according to 
equation \eqref{output_general}:

\begin{eqnarray}\label{output}
O_{V1}(x_1,x_2, \theta) &=&\int I(\xi_1,\xi_2)\psi
_{x_1,x_2,\theta, V1 }(\xi_1,\xi_2)d\xi_1 d\xi_2
\end{eqnarray}

\subsubsection{The subriemannian neurogeometry}

It is not surprising that there is a strong relation between the set of cell RPs and  their connectivity.  
Connectivity kernels have been measured by Bosking, who proved that they have an elongated shape  (\cite{bosking1997orientation}, see also figure 6 right), and that the strength of connectivity is maxima between cells with the same orientation. Due to this strong relation, we will define the metric on the space $R^2\times S^1$, 
starting from properties of the filters. We  introduce 1- form modeled on the expression of the filters  \eqref{filtri}:
$$\omega=  \sin(\theta)dx_1 - \cos(\theta)dx_2$$
and 
choose  left invariant vector fields, which lie in its kernel: 
$$X_1= \cos(\theta) \partial_x + \sin(\theta) \partial_y,\quad X_2= \partial_\theta.$$
The plane generated by these vector fields will be called Horizontal tangent bundle $HT$. 
Note that we have introduced only two vector fields in a 3D space, but the orthogonal direction can be recovered as a commutator: 
$$X_3= [X_1, X_2]= -\sin(\theta) \partial_x + \cos(\theta) \partial_y.$$
This condition, called the H\"ormander condition, allows to define a subriemannian structure, if we also choose a metric on $HT$. 
The metric is not unique and the simplest 
choice is the metric which makes the vector fields orthonormal. 
In other words 
for every vector $$v= a_1 X_1 + a_2 X_2
\text{ we will define norm of }  v \text{ the quantity }
||v||= \sqrt{a_1^2 + a_2^2}.$$
We will see in the next section that this choice 
can justify many perceptual phenomena, but other choices 
could be considered.

\subsubsection{Grouping and emergence of shapes}

Differential operators can be defined in terms of the vector fields $X_1$ and $X_2$, which play the same role of a 
derivative in the Euclidean setting. 
In particular when studying grouping we can consider the 
Fokker-Planck operator 
$$FP u = X_1 u + X_2^2u. $$
Due to the H\"ormander condition, it is known that this operator 
has a fundamental solution $\Gamma_{V1}$, which is strictly positive. 
However it is not symmetric in general, and in order to model horizontal connectivity we need to symmetrize it
$$K_{V1}(x_1, x_2 ,\theta, \xi_1, \xi_2 ,\eta) = 
\Gamma_{V1}(x_1, x_2 ,\theta, \xi_1, \xi_2 ,\eta) + 
\Gamma_{V1}( \xi_1, \xi_2 ,\eta, x_1, x_2 ,\theta).
$$
An isosurface of this kernel  is 
depicted in figure 7. 

\begin{figure}\label{fokker_kernel}
\begin{center}
\includegraphics[height=5 cm]{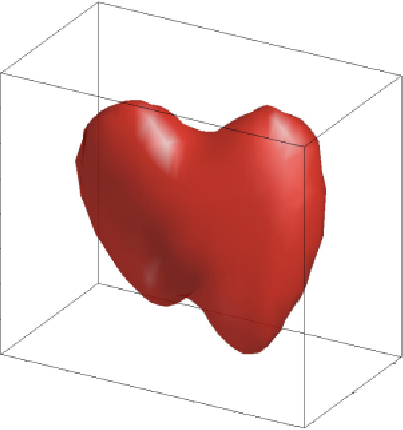}
\end{center}
\caption{A level set of the kernel $K_{V1}$.}
\end{figure}

To segment a stimulus into perceptual units we use then this kernel with the spectral 
technique introduced in \eqref{spectral_general}.
We can apply the eigenvalue equation to the operator associated to $K_{V1}$: 
$$K_{V1} * u = \lambda u$$
A few results have been depicted in figure 8. 
The first image on the left is the Kanizsa square. Due to the strong orientation properties of the kernel, collinear elements tend to be grouped. Hence the L-junctions 
pops up as the first eigenvector (highlighted in red). 
This can be considered the key element at the basis of perception of the square
(the second eigenvector, that is not visualized, corresponds to the 4 arcs of circle).
Note that the technique is invariant with respect to rotation, 
and the a L-junctions in Kanizsa diamond 
are detected with the same accuracy (second image in figure 8). 
In the last image we show that not only inducers of
classical geometrical object such as squares or diamonds 
are correctly detected, but also inducers of completely 
arbitrary shapes can emerge (third image in figure 8). Indeed the 
geometric process of grouping is based on the 
internal and global coherence of the elements in the image, 
not on a pre-existing list of possible shapes. 
As such it introduces a geometric intrinsic notion of shape based on the property of the cortex.

\begin{figure}\label{groupresult}
\begin{center}
\includegraphics[height=3.5 cm]{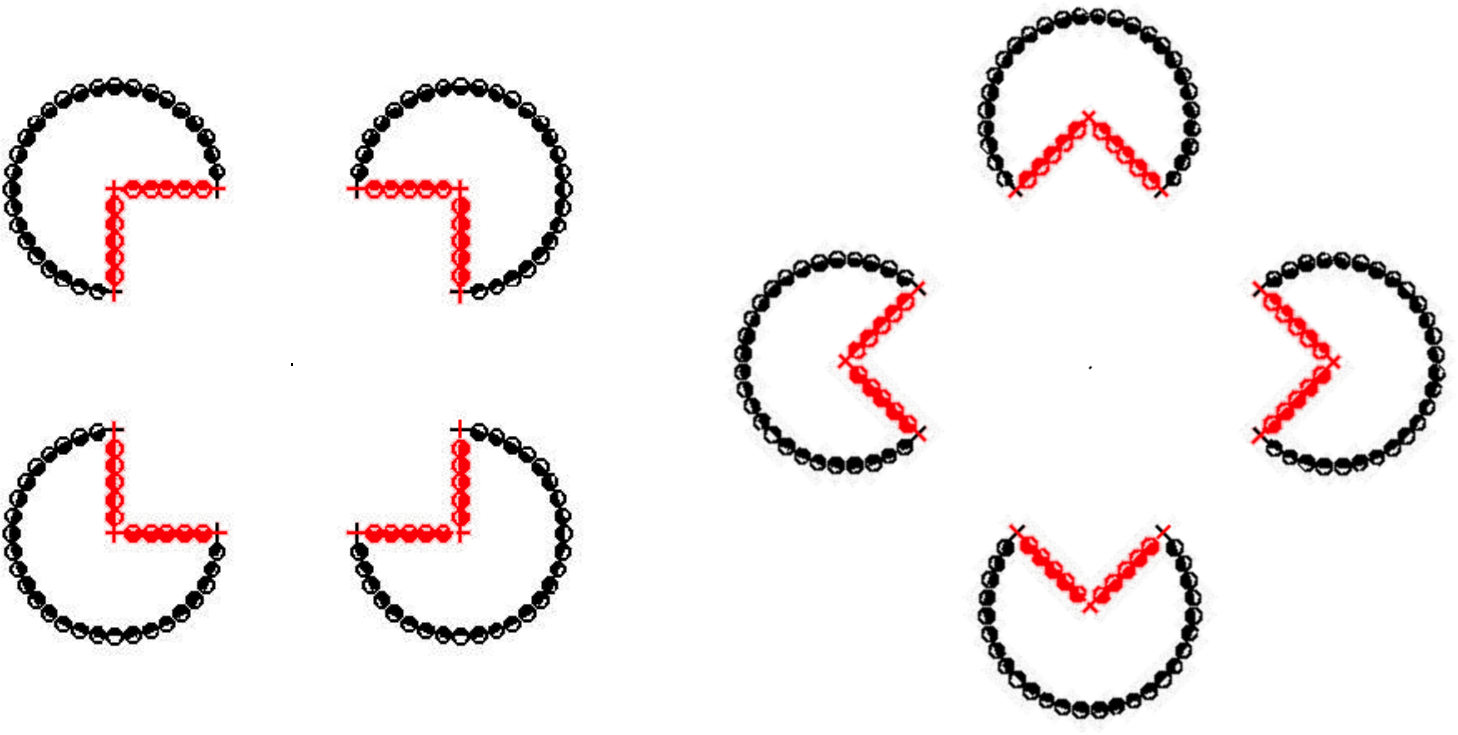}
\includegraphics[height=4.5 cm]{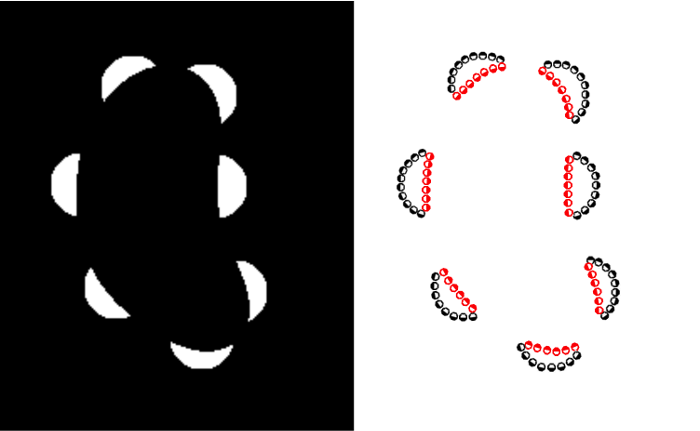}
\caption{The result of the grouping process applied from left to right to the Kanizsa square (as in \cite{favali17}) , to the Kanizsa diamond and to an other image.}
\end{center}
\end{figure}

\subsubsection{Completion in the orientation  geometry}

In order to project  on the image plane, we first  apply the equation  \eqref{max_general} restricted to this setting:
\begin{equation}
|O_I(x_1, x_2,  \theta_I(x_1, x_2))| = \max _\theta |O_I(x_1, x_2, \theta)|.
\end{equation} 

The cells are selective to the orientation since it can be proved that 
$\theta_I(x_1, x_2)$ is the direction of the level line of $I$ at the point $(x_1, x_2)$. As a consequence the gradient of $I$ can be approximated as follows
$$A= \nabla I \approx |O_I(x_1, x_2,  \theta_I(x_1, x_2))| (-\sin(\theta_I(x_1, x_2)), \cos(\theta_I(x_1, x_2))) .$$
Dividing by the norm of $A$ we obtain
$$\frac{A_1}{\sqrt{A_1^2 +A _2^2}} = -\sin(\theta_I)\quad 
\frac{A_2}{\sqrt{A_1^2 +A _2^2}} = \cos(\theta_I)
$$
and the 2D projection of the vector field 
 $X_1$ is expressed in terms of $A$ as 
$$
X_{1, A} = \frac{ A_2}{\sqrt{A_1^2 +A _2^2}} \partial_x  - \frac{ A_1}{\sqrt{A_1^2 +A _2^2}} \partial_y. 
$$

The  2D projection of the
Kernel defined in \eqref{kernel_projection},
becomes in this case 
$$
K_{\Pi}( (\xi_1, \xi_2), (x_1, x_2)) = K_{V1}(
(\xi_1,\xi_2, \theta_I (\xi_1,\xi_2), (x_1, x_2, \theta_I (x_1, x_2)). $$

\begin{figure}\label{2Dassociation}
\begin{center}
\includegraphics[height=3 cm]{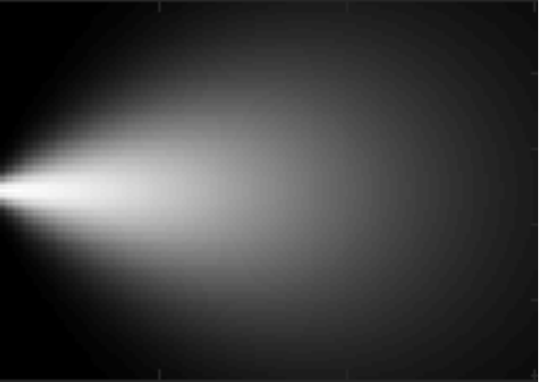}
\end{center}
\caption{The 2D projection $
K_{\Pi}$ of the kernel. $
K_{V1}$.}
\end{figure}

This kernel, depicted in figure 9, can be considered a good model of the association fields introduced by Field Hayes, Hess, in \cite{field1993contour}. 

Since the 
$K_{V1}$ expresses propagation in the direction $X_1$ and $X_2$, convolution with its projection 
$K_{\Pi}$ 
expresses propagation in the direction of the vector field 
$
X_{1, A} $. 
Propagation along the kernel can be expressed as minimizer of the functional 
\begin{equation}
\label{funct2}
L_2(A) = \int |X_{1, A} A|^2.
\end{equation}

\subsection{Combined contrast - orientation geometry}

In this section we will study the join action of LGN cells and cells in V1, taking into account 
feedforward, horizontal, and feedback connectivity. We propose a complete Lagrangian, sum of three terms. 
The first two have already been introduced:
the functional $L_1$ defined in \eqref{functional1} expresses the contrast variable 
considered as the analogous of a particle term in a gauge field functional and 
the functional $L_2$ defined in \eqref{funct2}
which describes a strongly anisotropic process and it is performed with respect to a subriemannian metric. 
Finally we will introduce  an interaction term coupling the two terms as in the usual Lagrangian field theories.

The last term describes the interaction between the particle $u(x,y)$ and the field $A(x,y)$. 

\begin{equation}\label{second}
L_3 =  \int |\nabla u(x,y) - A(x,y) |^2 dx dy.
\end{equation}

The resulting functional is then
\begin{equation}\label{functionalnorm}
\mathcal{L}=\int |\nabla u - \nabla I |^2 dx dy+\int |\nabla u - A |^2  dx dy + \int |X_{1 A}A |^2  dx dy.
\end{equation}

The Euler Lagrange equations of the functional (\ref{functionalnorm}) are obtained by variational calculus:
\begin{equation} \label{EL}
\Bigg\{\begin{matrix}  \Delta u =\frac{1}{2}( \Delta I + div (A)) \\ \Delta _A A =- \nabla u + A. \end{matrix}
\end{equation}
In order to solve the system we first solve the 
particle  equation, assuming that $A=0$. Then, a first estimate 
$u_1$ of $u$ is obtained   as a solution of  the first equation
\begin{equation}  
\Delta u_1 =\frac{1}{2}( \Delta I)  \end{equation}
The function $u_1$  select boundaries (see figure 10(a)) 
Then we apply the grouping algorithm and find the inducers of the square (see figure 10(b)). With this estimated second member and the condition $A_0=0$ we find the solution of the equation for the vector $A$: 
\begin{equation}  
\Delta_A A =- \nabla u_1 
\end{equation}
The diverge of the vector solution is depicted in fig 10(c).
 Finally we recover the value of $u$ using as forcing term  $\Delta I + div(A)$, 
\begin{equation}  
\Delta u =\frac{1}{2}( \Delta I + div(A))  \end{equation}
applying to the function $u$ the mean as in \eqref{br} we obtain 
$$b= (I+u)/2,$$
which is the last image represented in figure 10.

\begin{figure}[H]
\begin{center}\label{Kan0}
\includegraphics[height=3.5 cm]{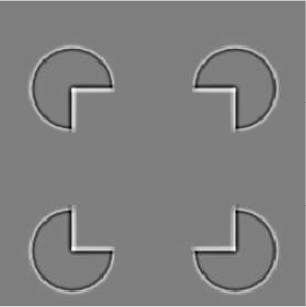}
\includegraphics[height=3.5 cm]{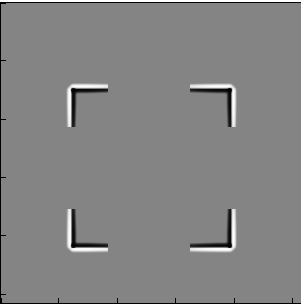}
\includegraphics[height=3.5 cm]{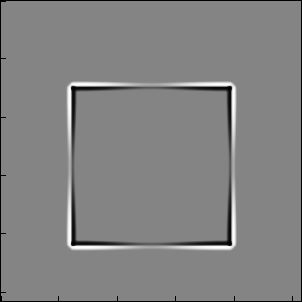}
\includegraphics[height=3.5 cm]{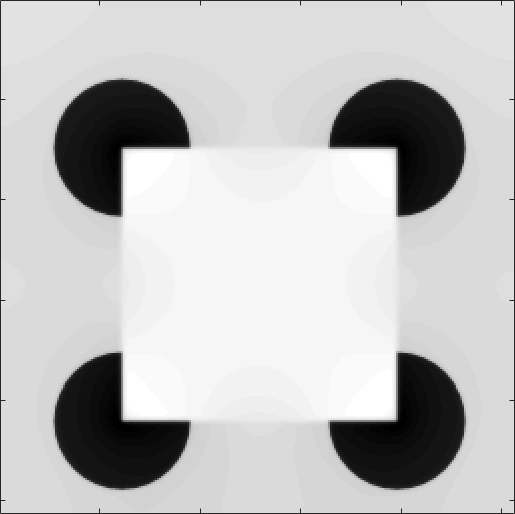}
\end{center}
\caption{From left to right: a) The laplacian $u_1$ of the Kanizsa square, b) the inducers found with the grouping algorithm, c) the vector $A$, which completes the square d) the final value of the function $b$}
\end{figure}

The same combined algorithm is also applied to the a Kanizsa diamond and to a non symmetric shape in figure 11. 

\begin{figure}[H]\label{combined}
\begin{center}
\includegraphics[height=3.5 cm]{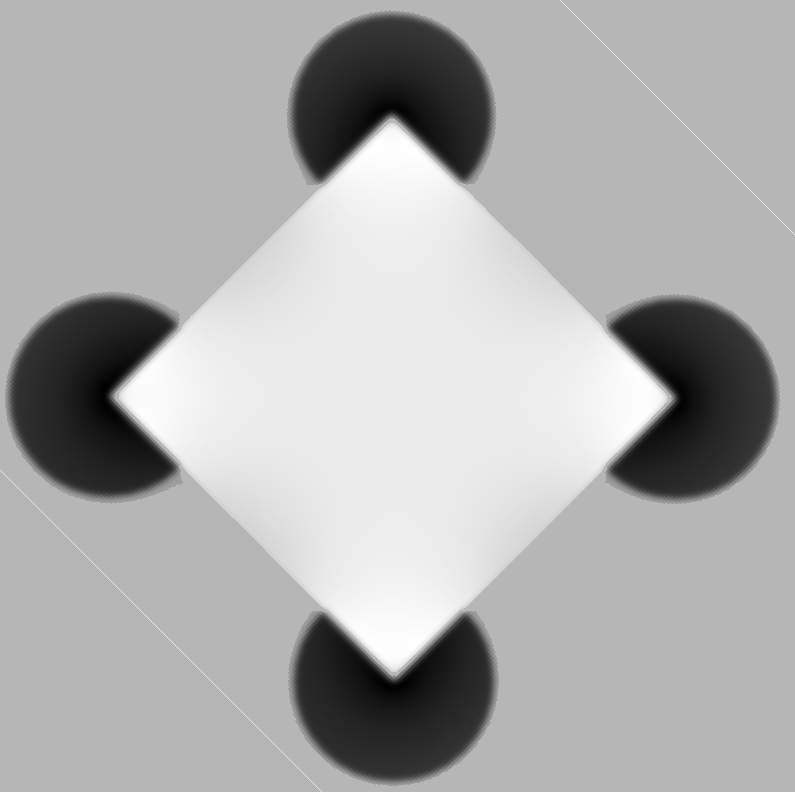}
\includegraphics[height=4 cm]{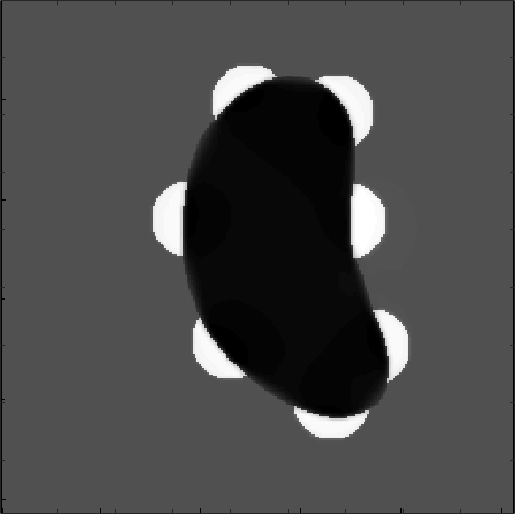}
\caption{The combined contrast/orientation algorithm is applied to the Kanisza diamond and to an other image.}
\end{center}
\end{figure}

Due to the fact that the model is invariant by rotation symmetry the completion of the Kanizsa square is completely 
equivalent to the completion of the Kanizsa diamond up to a global rotation. We will question in the following this rotation invariance based both on perceptual evidence and image statistics results.

\subsection{The model applied to other features}
The previous model can be extended to the description of 
families of cell sensitive to other features, 
as for example movement. This has been studied in \cite{cocci2015cortical}, where grouping and 
completion has been achieved in a five-dimensional space,
depending on position $(x_1, x_2)$, orientation, time and apparent velocity. 
We also refer to \cite{Franceschiello} 
 where other classical illusion only dependent on orientation have been considered 
 (Hering, Wundt, Ehrenstein, Zollner illusions). We expect that better results could be obtained with the algorithm we present here, able to combine the grouping effect with the 
orientation selectivity. 

An other feature which could be considered is 
the disparity feature, which allows 
reconstruction a 3D image from 2 flat projections 
on the retina of the eyes. With our approach these aspects of perception 
of images have to be developed in the group $SE(3)$ of rigid 
motion of the 3D space. 
This phenomena is probably related to the distal properties of 
brightness perception. Indeed according to \cite{Todorovic} 
a flat image exhibiting various reflectance distributions,
observed under uniform illumination, can be interpreted as 
a scene with different level of illumination, giving the impression of a 3D object, 
and giving rise to illusions of perception of brightness.

\section{Anysotropic aspects of the geometry induced by learned kernels} 

Studies of visual acuity in humans were among
the first investigations to uncover preferences
for vertical and horizontal stimuli. Emsley
(192S) noticed acuity differences among subjects
asked to resolve line gratings. Maximal
acuity occurred when the gratings were in
horizontal or vertical orientations.
We will see that this anisotropy can be explained in terms of kernel learned by natural images.

\subsection{Statistics of images} 

 An intriguing hypothesis is that association fields induced by horizontal connectivity have been learned by the history of visual stimuli that a subject has perceived.  In this case the specific connectivity  pattern would result from some statistical property of natural images.

A specific study to assess the existence of this relation has been proposed in  \cite{sanguinetti}. Research has been focused  on the statistics of edges in natural images and particularly in the statistics of co-occurence  of couples of edges taking into account its relative position and orientation. 
The statistics has been estimated analyzing a number of natural images 
from the image database:
http://hlab.phys.rug.nl/imlib/index.html, which has been many
times used in literature to compute natural image statistics
It consists on 4000 high-quality gray scale digital images,
1536 $\times$ 1024 pixel and 12 bits in depth.
Images have been preprocessed by linear filtering with a set of oriented edge detection kernels (Gabor filters)  and performing non maximal suppression. A list of pixels corresponding to edges with their respective orientations has been obtained by thresholding and binarization.  If we fix the orientations  $\theta_c$ 
and $\theta_p$, a $2\times 2$ histogram can be computed 
counting how many edges have orientation $\theta_c$
at the central pixel 
and orientation $\theta_p$ at the position $(\Delta x, \Delta y)$. An example is depicted in figure 12, left. 
If we consider all possible orientations $\theta_c$  of the central pixel and 
$\theta_p$ of the other pixels, we obtain a 
four dimensional histogram $(\Delta x, \Delta y,\theta_c, \theta_p)$ (see fig. 12, right).

\begin{figure}[H]
\centering
\includegraphics[height=3 cm]{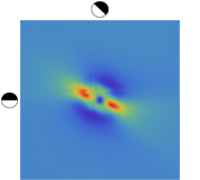}\;\;\;\;\;\;
\includegraphics[height=7 cm]{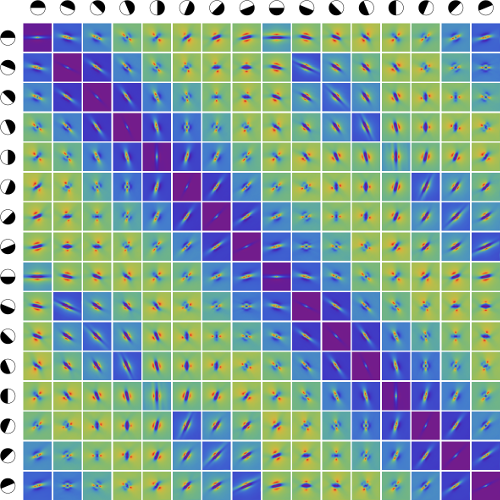}
\caption Left: a 2D histogram of occurrence of edges. The color at the point 
 $(\Delta x, \Delta y)$ represents the number of edges with orientation  $\theta_c$ (diagonal)
at the central point and orientation $\theta_p$ (horizontal)  at the  point  $(\Delta x, \Delta y)$. The red colour represents the highest values, and blue the lowest. Right: 
the four dimensional histogram. In each row the direction  $\theta_c$ of the central pixel is fixed, while the final orientation 
 $\theta_p$ takes  different values, and the corresponding 2D histogram is represented.

\end{figure}

\subsection{Anysotropic Connectivity kernels }

In \cite{sanguinetti} a 3D histogram  $(\Delta x, \Delta y,\Delta \theta)$ is obtained where the third coordinate is the relative orientation $\Delta \theta = \theta_p-\theta_c$.
This  kernel,  invariant by rotation and translation was 
compared with the one obtained 
with the Lie group theory.

Here we keep now completely independent any orientation 
in order to study possible anisotropy of the statistic in the 
different directions. 

We focus in particular on the 
histogram with the central orientation horizontal and the one with the central orientation diagonal. 
This means that we consider separately the first and the third row of the histogram. 

\begin{figure}[h]
\centering
\includegraphics[height=3 cm]{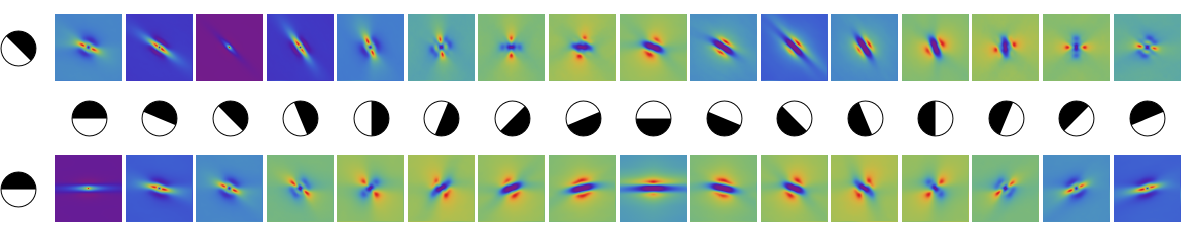}
\caption{The first and the third row of the histogram representing the statistics of edge co-occurrence with respect to a central orientation respectively horizontal or diagonal}
\end{figure}

We have selected two histograms each one  constituted by 16 
$128\times 128$ images. 
Hence each of them  can be considered a 
$128\times 128\times 16$
 array, discretization of 
 a 3D kernel. We will denote 
the two histograms respectively associated with the horizontal or diagonal central orientation as follows:
$$K_{hor}: R^2 \times S^1 \rightarrow R \quad \text{  and  }\quad 
K_{diag}: R^2 \times S^1 \rightarrow R.$$
We visualized each of them 
in figure 13 by slices and figure 14 in 3D: 
it is clear that they are different, even though they have some similarity with the 
3D kernel in figure 7. 
In particular if we assume that cortical connectivity is modulated by the geometry of the perceived images, then the connectivity kernel will be  different  for different directions. 

\begin{figure}[h]\label{histo_hd}
\centering
\includegraphics[height=5.5 cm]{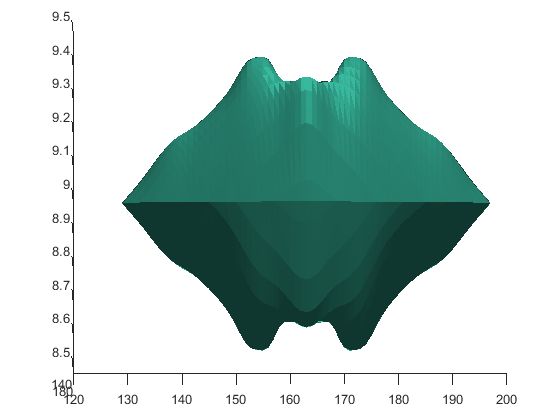}
\includegraphics[height=5.5 cm]{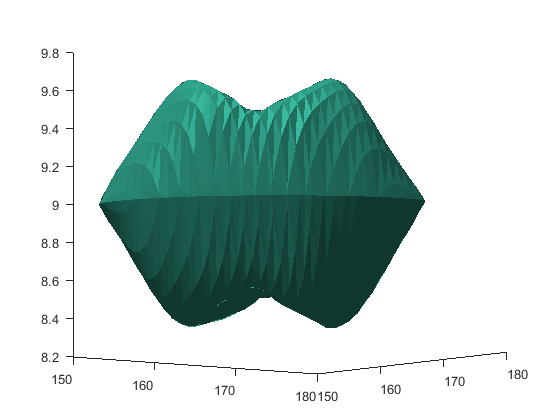}
\caption{The histograms $K_{hor}$ and $K_{diag}$ of edge co-occurence related to the horizontal and diagonal direction are function defined on a 3D space with real values. For a fixed value of $c$ we depict here a level surface$K_{hor}(x,y,\theta) = c$ (left) and  $K_{hor}(x,y,\theta) = c$ (right). These level surfaces are represented as surfaces in the 3D domain of the kernels.}
\end{figure}

\subsection{Anisotropic association fields}

We can project the kernels $K_{hor}$ and $K_{diag}$ 
on the 2D plane, in order to understand their properties. 
We obtain respectively 
$$K_{\Pi, hor}(x_1, x_2, \xi_1, \xi_2) = \max_\theta K_{hor}(0,0, x_1, x_2, \theta)  \quad K_{\Pi, diag}(x_1, x_2, \xi_1, \xi_2) = \max_\theta K_{diag}(0,0, x_1, x_2, \theta) $$
The two kernels are depicted in figure 15. We can see that they have completely different properties. The horizontal one takes higher values, and has a longer action. Moving from the diagonal the values change very rapidly. Hence the kernel has a very clear ability to discriminate the horizontal direction with respect to any other one. 

The diagonal one takes lower  values, and has a shorter action, and does not change very much far from the diagonal. 
As a consequence the kernel assume comparable values for diagonal and slightly off diagonal segments. 

\begin{figure}[H]
\begin{center}\label{ass}
\includegraphics[height=3.5 cm]{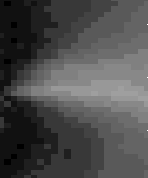}
\includegraphics[height=3.5 cm]{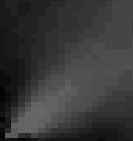}\,
	\caption{A model of association field inspired by statistic of images. The horizontal one (left) and the diagonal one (right) have completely different geometrical properties.} 
\end{center}
\end{figure}

\subsection{Difference in horizontal versus diagonal orientation detection} 
Here we would like to related 
the geometry of the non isotropic association 
fields we have found and the
difference in orientation detection for stimuli
aligned in horizontal as compared to stimuli in oblique
orientations. 

We start with a few horizontal segments, 
which can be evaluated with the horizontal association field.
Since the intensity of the kernel is very high, 
it takes the same values on aligned horizontal segment 
(see figure 16 above left), and keep the same value even if these are taken apart (see figure 16 above middle). 
On the other side the values change very rapidly while moving 
from the horizontal, hence the kernel takes values very different 
on non aligned segments, leading to a a very high discrimination (see figure 16 above right) in accordance with psycophysical results. 

\begin{figure}[H]
\begin{center}\label{lines}
\includegraphics[height=4 cm]{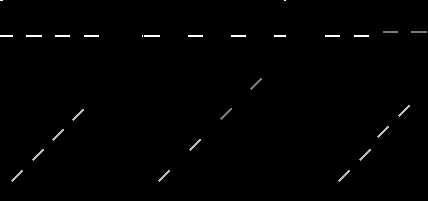}\;\;
\caption{The value of the association kernel on different configurations of horizontal an diagonal segments. }
\end{center}
\end{figure}

When we repeat the experiment on diagonal segments the situation is very different. The kernel tends to group aligned diagonal  segments (see figure 16 bottom),
but since the intensity of the kernel decreases rapidly with 
the distance,  the  kernel is not able to group distant segments,
even if they are diagonal. 
Finally the values of the diagonal kernel change very slowly  while moving 
from the diagonal, hence the kernel takes the same 
values on diagonal misaligned segments, leading to a low  discrimination, in accordance with the so called oblique effect.

\subsection{Oblique effect in the Kanizsa square} 

The oblique effect was studied in Kanizsa square with misaligned edges in \cite{Bross}. The problem refers to the ability to perceive stimuli lying in the horizontal orientation with greater efficiency than stimuli situated in any other orientation. It was proved that 
the  stimuli in the oblique orientation is perceived as square for greater degrees of misalignment than those in the cardinal axis orientation (see figure 3). 

Here we simulate the experiment with our model and for misalignment at 0, 6, and 12 degrees for the Kanizsa squares and diamonds (see figure 17). 
We first express the propagation with the two different kernels $K_{\Pi hor}$ and $K_{\Pi diag}$ in terms of fundamental solution of differential equations. 
The horizontal kernel is strongly directional, so that it will be represented as a pure subriemannian operator. Consequently the Lagrangian operator will be exactly the  
one introduced in \eqref{functionalnorm}: 
\begin{equation}
\mathcal{L}_{hor}=\int |\nabla u - \nabla I |^2 dx dy+\int |\nabla u - A |^2  dx dy + \int |X_{1 A}A |^2  dx dy.
\end{equation}
The diagonal kernel is weaker and less directional, so that we will add a Riemannian coefficient to better model it. Consequently the Lagrangian operator will be: 
\begin{equation}
\mathcal{L}_{diag}=\int |\nabla u - \nabla I |^2 dx dy+\int |\nabla u - A |^2  dx dy   + \frac{1}{2}\int |X_{1 A}A |^2  dx dy + \epsilon \int |X_{3 A}A |^2  dx dy .
\end{equation}

The minimizers of the two functionals are computed and results are presented in figure 18: as in the perceptual experiment, 
both the Kanizsa square and diamond
are completed if there is no misalignment (left image). In presence of a misalignment of 6 degrees, 
due to the strongly directional properties of the kernel, the horizontal Kanizsa square is not reconstructed, while the strongly diffusive effect of the diagonal kernel does not allow to appreciate the misalignment, and the diamond is much better completed in accordance with experimental results (middle image).
Finally for a misalignment of 12 degrees, neither the square nor the diamond are reconstructed (right image).


\begin{figure}[H]
\begin{center}
\includegraphics[height=4 cm]{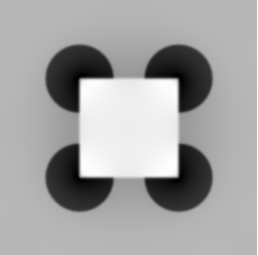}
\includegraphics[height=4 cm]{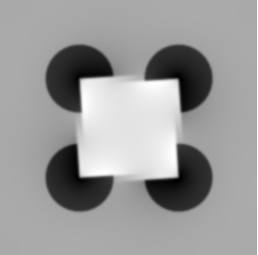}
\includegraphics[height=4 cm]{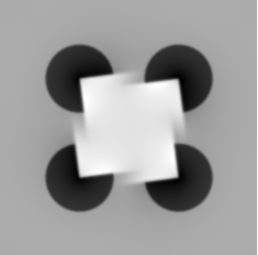}
\includegraphics[height=4 cm]{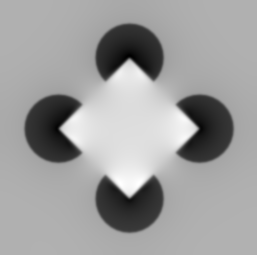}
\includegraphics[height=4 cm]{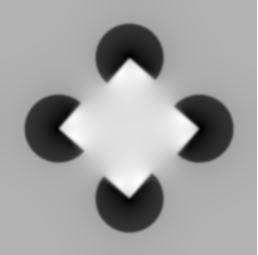}
\includegraphics[height=4 cm]{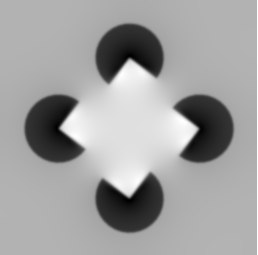}
\end{center}
\caption{The reconstruction made by our model. In good agreement with the experiments, with edge misalignment of 0, degrees both the square and the diamond are reconstructed (left images), for misalignment of 6 degree the square is not reconstructed, while the diamond is reconstructed (middle images). Finally for 12 degrees neither the square not the diamond are reconstructed (right images).}
\end{figure}

\section{Conclusion} 
We presented here a general model, which unifies a number of models of the neurogeometry of perception. 
Every  family of cells,  is parametrized with coordinates $R^2 \times F$, where $R^2$ represents the position on the retinal plane and $F$ the extracted feature. The cortical  connectivity 
is expressed via a kernel $K_F$, whose eigenvectors can describe the emerge of perceptual units \cite{sarti2015constitution} and represent the distribution of neural activity in un the primary cortex. Cortical  connectivity kernels, projected on the 2D plane induces a geometry of perception on the visual plane. Propagation with these kernels is responsible for perceptual completion. The model is first applied to contrast and orientation perception using kernels invariant with respect to suitable Lie group laws following the presentation of \cite{gaugecortex}. These models are able to justify modal completion as for example the Kanzsa square. However they discard the ability of the visual system to perceive stimuli lying in the horizontal orientation with greater efficiency than in any other orientation. In order to take into account this aspect, we applied the model with non isotropic kernels learned by statistical of images, recovering anisotropy in the geometry of perception in different orientations.
A more radical heterogeneity able to account for different geometries locally defined will be considered in the future. A preliminary study has been presented in 
 \cite{SCP}.

\end{document}